\documentclass[letterpaper, 10 pt, conference]{ieeeconf}  

\IEEEoverridecommandlockouts                              

\usepackage{amsmath} 

\usepackage{graphicx}
\usepackage{makecell}
\usepackage{multirow}
\usepackage{booktabs}
\usepackage{xspace}
\usepackage{xcolor}
\let\labelindent\relax
\usepackage{enumitem}

\makeatletter
\let\NAT@parse\undefined
\makeatother
\usepackage[numbers,sort&compress]{natbib}

\usepackage{hyperref}
\definecolor{lightblue}{rgb}{0.68, 0.85, 0.9}
\definecolor{darkblue}{rgb}{0.48, 0.65, 0.7}
\definecolor{lightgray}{gray}{0.9}
\definecolor{citecolor}{HTML}{0071BC}
\hypersetup{colorlinks,linkcolor={red},citecolor={citecolor}}

\newcommand{\alias}{ReBot\xspace}

\title{\LARGE \bf
\alias: Scaling Robot Learning with \\ Real-to-Sim-to-Real Robotic Video Synthesis
}

\author{Yu Fang$^{1}$, Yue Yang$^{1}$, Xinghao Zhu$^{2}$, Kaiyuan Zheng$^{3}$, Gedas Bertasius$^{1}$, Daniel Szafir$^{1}$, Mingyu Ding$^{1}$
\thanks{$^{1}$Yu Fang, Yue Yang, Gedas Bertasius, Daniel Szafir, and Mingyu Ding are with Department of Computer Science, University of North Carolina at Chapel Hill, 201 S Columbia St, Chapel Hill, NC 27599, USA. {\tt\small \{yufang, yygx, gedas, dszafir, md\}@cs.unc.edu}}%
\thanks{$^{2}$ Xinghao Zhu is with Robotics and AI Institute, 145 Broadway, Cambridge, MA 02142, USA. {\tt\small xizhu@rai-inst.com}}
\thanks{$^{3}$ Kaiyuan Zheng is with Department of Electrical and Computer Engineering, University of Washington, 1410 NE Campus Parkway, Seattle, WA 98195, USA. {\tt\small kaiyuan5@uw.edu}}
}

\begin{document}

\maketitle
\thispagestyle{empty}
\pagestyle{empty}

\begin{abstract}
Vision-language-action (VLA) models present a promising paradigm by training policies directly on real robot datasets like Open X-Embodiment.
However, the high cost of real-world data collection hinders further data scaling, thereby restricting the generalizability of VLAs.
In this paper, we introduce \alias, a novel real-to-sim-to-real approach for scaling real robot datasets and adapting VLA models to target domains, which is the last-mile deployment challenge in robot manipulation.
Specifically, \alias replays real-world robot trajectories in simulation to diversify manipulated objects (real-to-sim), and integrates the simulated movements with inpainted real-world background to synthesize physically realistic and temporally consistent robot videos (sim-to-real).
Our approach has several advantages: 1) it enjoys the benefit of real data to minimize the sim-to-real gap; 2) it leverages the scalability of simulation; and 3) it can generalize a pretrained VLA to a target domain with fully automated data pipelines.
Extensive experiments in both simulation and real-world environments show that \alias significantly enhances the performance and robustness of VLAs.
For example, in SimplerEnv with the WidowX robot, \alias improved the in-domain performance of Octo by 7.2\% and OpenVLA by 21.8\%, and out-of-domain generalization by 19.9\% and 9.4\%, respectively.
For real-world evaluation with a Franka robot, \alias increased the success rates of Octo by 17\% and OpenVLA by 20\%.
More information can be found at our \href{https://yuffish.github.io/rebot/}{project page}.

\end{abstract}

\section{Introduction}
Large-scale real robot datasets have demonstrated their significant contribution to the rapid advances of robot learning~\cite{brohan2022rt,brohan2023rt,padalkar2023open}, enabling vision-language-action (VLA) models to learn across various tasks, environments, and embodiments.
Despite these achievements, VLAs still face challenges in effectively generalizing to new scenarios, spurring the need for scaling data to enhance their performance in new target domains.
However, collecting large-scale real robot datasets is very costly and often demands extensive effort and resources, \emph{e.g.}, robots and human teleoperators, which significantly limits the availability and scalability~\cite{o2023open, khazatsky2024droid}.
On the other hand, simulated datasets are more accessible and cost-effective alternatives, as they can be generated in simulation environments without real-world setups~\cite{kolve2017ai2, mu2021maniskill, gu2023maniskill2, wang2023robogen,liu2023libero,nasiriany2024robocasa}. 
Unfortunately, the sim-to-real gap in both the action space and the observation space hinders robot policies from generalizing to real-world applications~\cite{zhao2020sim, muratore2022robot}, limiting the effectiveness of simulated data for advancing VLAs.

\begin{figure}[t]
  \centering
   \includegraphics[width=0.99\linewidth]{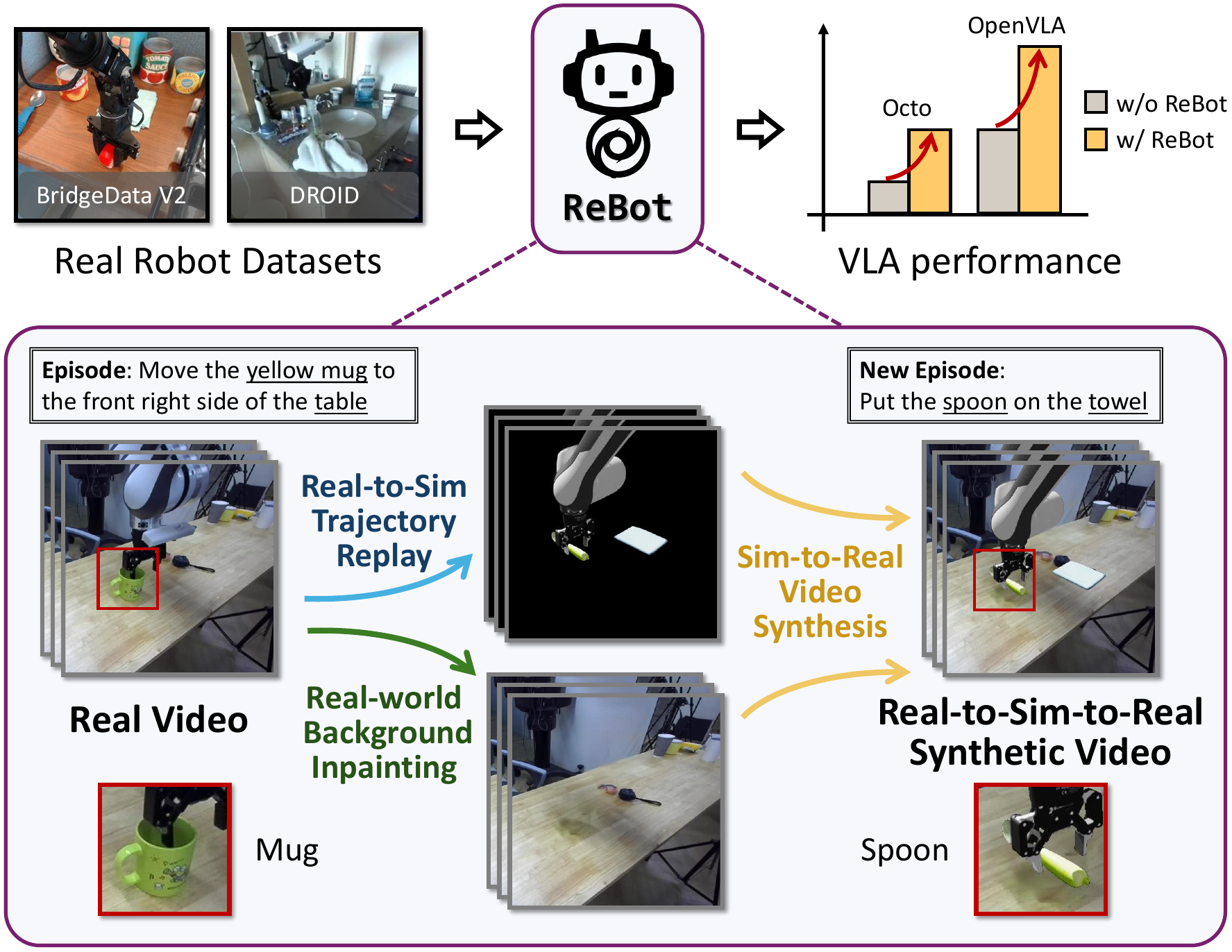}
   \vspace{-15pt}
   \caption{
   \textbf{An overview of \alias.}
   We propose \alias, a novel real-to-sim-to-real approach for scaling real robot datasets.
   ReBot replays real-world robot trajectories in a simulation environment to diversify manipulated objects (real-to-sim), and integrates the simulated movements with inpainted real-world background to produce realistic synthetic videos (sim-to-real), effectively adapting VLA models to target domains.
   }
   \label{fig:teaser}
   \vspace{-1em}
\end{figure}

To tackle these challenges, a straightforward strategy for scaling robot learning is generating synthetic robot videos from real robot datasets.
With the rapid development of foundation models in computer vision and generative AI, researchers have introduced generative models for synthetic robot video generation~\cite{mandi2022cacti,zhou2024robodreamer,du2024learning}.
For example, methods~\cite{chen2023genaug,yu2023scaling, chen2024roviaug} have leveraged text-to-image inpainting to scale real robotic images to diverse scenarios.
However, they typically face the issue of AI-generated artifacts such as visible imperfections or inconsistent textures, failing to produce physically realistic and temporally consistent robot videos.
Such distortions introduce new domain gaps, making it difficult for VLAs to learn stable and continuous robot actions while raising reliability concerns.
Additionally, generated images may not adhere precisely to instruction conditions, limiting the effectiveness of such methods in adapting VLAs to specific target domains, leaving the last-mile deployment challenge in robot manipulation unresolved.

To mitigate these issues, we propose \alias, a novel real-to-sim-to-real approach for scaling real robot datasets and adapting VLA models to target domains.
Our key insight is to replay real-world robot trajectories in simulation to diversify manipulated objects (real-to-sim), and integrate the simulated movements with inpainted real-world background (sim-to-real) to synthesize physically realistic and temporally consistent robot videos.
Notably, \alias combines the advantages of both sim and real, \emph{i.e.}, leveraging the scalability of simulation, while minimizing the sim-to-real gap by grounding both the action and observation spaces from real robot data.
Particularly, in contrast to generation-based scaling approaches, \alias ensures physical realism and temporal consistency, and enables effective adaptation of VLA models to target domains.

Specifically, as shown in Fig.~\ref{fig:teaser}, ReBot includes three key components:
1) Real-to-Sim Trajectory Replay. For each real-world episode, we automatically set up digital twins in a simulation environment, and replay the real-world robot trajectory to obtain simulated movements for manipulating new objects.
We validate the scalability of our approach by demonstrating that real-world trajectories can be successfully reused to manipulate different shapes of objects in simulation.
2) Real-world Background Inpainting. To obtain task-agnostic real-world background for video synthesis, we introduce an automated inpainting module with GroundedSAM2~\cite{ren2024grounded} to segment and track the robot and object (\emph{i.e.}, task-specific elements) in original real-world videos, and remove them with ProPainter~\cite{zhou2023propainter}.
3)  Sim-to-Real Video Synthesis. We eventually integrate simulated movements with task-agnostic real-world background, producing synthetic videos with realistic physics and excellent temporal consistency.

In summary, our key contributions are three-fold.
\begin{itemize}[leftmargin=*]
    \item We introduce \alias, which, to our knowledge, is the first real-to-sim-to-real approach for scaling real robot datasets and adapting VLA models to target domains, addressing the last-mile deployment challenge in robot manipulation.

    \item \alias combines the advantages of both sim and real, \emph{i.e.}, leveraging the scalability of simulation, while minimizing the sim-to-real gap by grounding both the action and observation spaces from real robot data. Notably, \alias is fully automated and requires no manual intervention.
    
    \item Extensive evaluations confirm \alias's effectiveness in both simulation and real-world settings, \emph{e.g.}, it improves OpenVLA’s in-domain and generalization performance by 21.8\% and 9.4\% on SimplerEnv and achieves a 20\% gain in real-world tasks, significantly outperforming prior state-of-the-art ROSIE~\cite{yu2023scaling}.

\end{itemize}

\section{Related Work}

\noindent\textbf{Scaling Robot Learning.}
Although many research institutes have collaborated to construct large-scale real robot datasets~\cite{o2023open, khazatsky2024droid}, data scale remains a fundamental bottleneck for VLA models.
To address this issue, recent works have explored three primary strategies: 1) collecting data in real-world environments, 2) collecting data in simulation environments, and 3) scaling real robot datasets with generative models.
Real robot datasets can be acquired using various methods, including kinesthetic teaching~\cite{ravichandar2020recent, yang2024enhancing}, teleoperation~\cite{mandlekar2019scaling, ebert2021bridge, jang2022bc, khazatsky2024droid}, or mixed reality devices~\cite{whitney2019comparing, yang2024arcade}, and have significantly contributed to the recent progress in VLA models~\cite{octo_2023, kim2024openvla}.
However, collecting large-scale real robot datasets demands extensive resources, making it highly challenging to scale across diverse environments and tasks.
This limitation hinders the generalization performance of VLA models.
On the other hand, simulated datasets offer a more scalable alternative.
Well-developed simulation platforms~\cite{savva2019habitat, shridhar2020alfred, xiang2020sapien, mittal2023orbit} facilitate rapid data collection in controlled environments without the high cost of real-world experiments.
Unfortunately, these datasets often introduce significant sim-to-real gap~\cite{zhao2020sim}, limiting their effectiveness in real-world applications.
Notably, recent works have explored generative models to scale real robot datasets~\cite{chen2023genaug,yu2023scaling, chen2024roviaug}.
Yet, these approaches often struggle to provide physically realistic and temporal consistent robot videos, making them unreliable and ineffective for developing VLA models.
In this paper, we propose a real-to-sim-to-real approach for scaling real robot datasets, offering a novel solution for these longstanding challenges.

\noindent\textbf{Real-to-sim and Sim-to-real.}
Real-to-sim and sim-to-real strategies have been explored in many applications in robotics~\cite{muratore2022robot, wang2023real2sim2real, torne2024reconciling, mu2024robotwin, li2024robogsimreal2sim2realroboticgaussian}.
Notably, recent work has leveraged real-to-sim-to-real strategy to develop simulated evaluation platforms for robotics~\cite{li2024evaluating}, demonstrating a strong correlation with real-world robot evaluations.
These studies highlight the significant potential of real-to-sim-to-real approaches in bridging the gap between simulation and real-world environments.
However, existing methods often face scalability challenges due to limited scene and object diversity, primarily due to the substantial manual effort for constructing digital twins in simulation environments~\cite{mu2024robotwin, li2024evaluating}.
In this paper, we explore a new application of this strategy, \emph{i.e.}, for scaling real robot datasets, enabling realistic robotic video generation without manual intervention.

\begin{figure*}[t]
  \centering
   \includegraphics[width=1.0\linewidth]{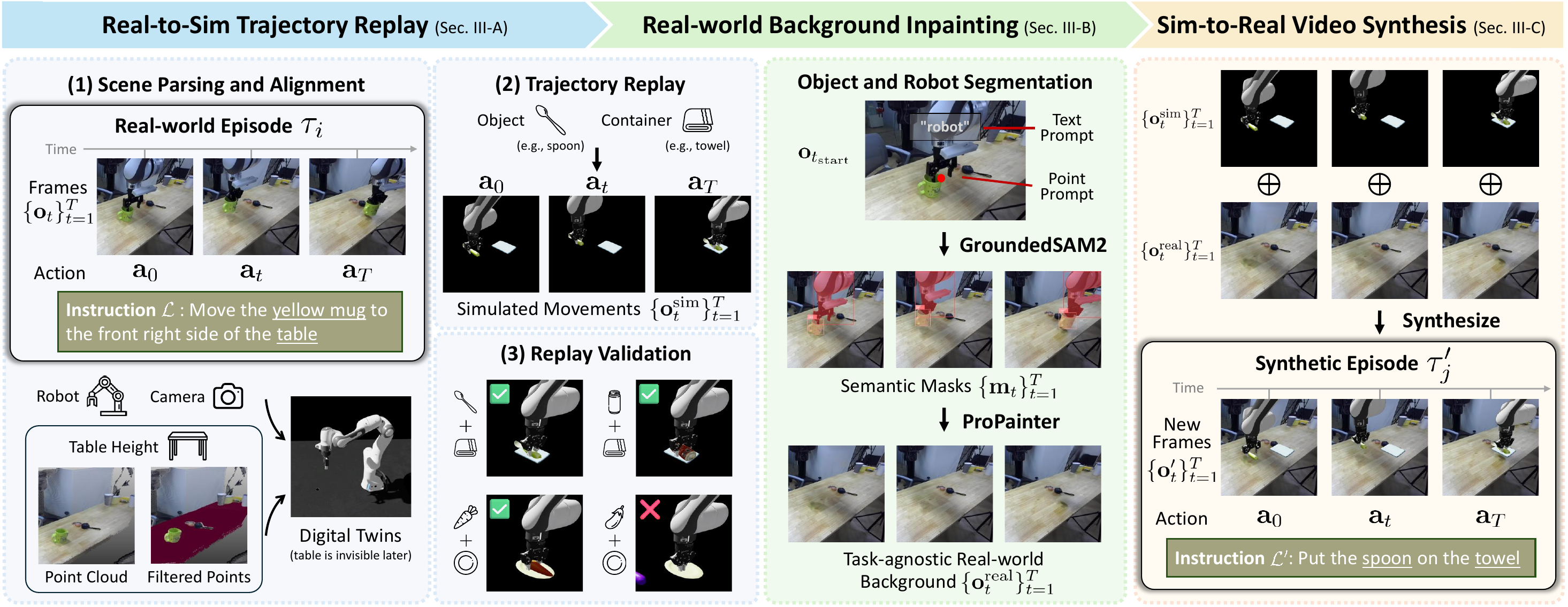}
   \vspace{-1.5em}
   \caption{\textbf{An overview of our framework.} ReBot includes three key components: 
   \textbf{A) Real-to-Sim Trajectory Replay:} For each real-world episode, we automatically set up digital twins and replay the real-world trajectory to obtain simulated movements for manipulating new objects. Each trajectory can be reused for different objects. 
   \textbf{B) Real-world Background Inpainting:} To obtain task-agnostic real-world background for video synthesis, we introduce an automated inpainting module to segment and remove the robot and object from the original real-world video.
   \textbf{C) Sim-to-Real Video Synthesis:} We eventually integrate simulated movements with task-agnostic real-world background to produce synthetic videos.
   ReBot is fully automated and requires no manual intervention.
   }
   \label{fig:method}
   \vspace{-1em}
\end{figure*}

\section{Method}
In this paper, we propose a novel real-to-sim-to-real approach for scaling real robot datasets.
We define a real robot dataset as $\mathcal{D} = \{\mathbf{\tau}_{i}\}_{i=1}^M$, where $M$ episodes are represented as $\mathbf{\tau}_i= \{\mathbf{o}_{t}, \mathbf{a}_{t}, \mathcal{L}\}_{t=1}^T$.
Here, $t$ denotes the timestep, $\mathbf{o}_t$ is the video frame, $\mathbf{a}_t$ is the action, $\mathcal{L}$ is the language instruction.
Our goal is to produce new synthetic episodes $\mathbf{\tau}'_j = \{\mathbf{o}'_t, \mathbf{a}_t, \mathcal{L}'\}_{t=1}^T$ based on $\mathbf{\tau}_i$, to build a synthetic dataset $\mathcal{D}' = \{\mathbf{\tau}'_{j}\}_{j=1}^N$ for adapting VLA models to target domains.
As illustrated in Fig.~\ref{fig:method}, ReBot has three key steps:
A) Real-to-Sim Trajectory Replay to obtain simulated movements $\{\mathbf{o}_{t}^{\text{sim}}\}_{t=1}^T$ in a simulation environment (Sec.~\ref{sec:real-to-sim});
B) Real-world Background Inpainting on video frame $\{\mathbf{o}_{t}\}_{t=1}^T$ to obtain task-agnostic real-world background $\{\mathbf{o}_{t}^{\text{real}}\}_{t=1}^T$ (Sec.~\ref{sec:real-world});
and eventually C) Sim-to-Real Video Synthesis to obtain new frame $\{\mathbf{o}_{t}'\}_{t=1}^T$~(Sec.~\ref{sec:sim-to-real}).

\subsection{Real-to-Sim Trajectory Replay}
\label{sec:real-to-sim}
The real-to-sim process involves: 1) Creating spatially aligned digital twins of the scene in the simulation environment, 2) Replaying real-world robot trajectory to produce simulated robot movements $\{\mathbf{o}_{t}^{\text{sim}}\}_{t=1}^T$, 3) Validating each replayed trajectory to ensure successful object manipulation.

\noindent\textbf{Scene Parsing and Alignment.}
To ensure faithful trajectory replay, we construct digital twins of the robot, cameras, and table, and align them to the initial video frame $\mathbf{o}_{1}$.
The prototypes of the robot and cameras are prepared ahead, only requiring pose adjustments to complete their setup.
To determine the table height, we acquire the metric depth from the initial video frame $\mathbf{o}_{1}$ and create a point cloud of the scene.
Using GroundingDINO~\cite{liu2023grounding}, we automatically segment the table with the text prompt (``table"), and extract the subset of the point cloud after removing outliers using the interquartile range.
We eventually set the average height of the filtered points as the table height.

\begin{figure*}[t]
  \centering
  \includegraphics[width=1.0\linewidth]{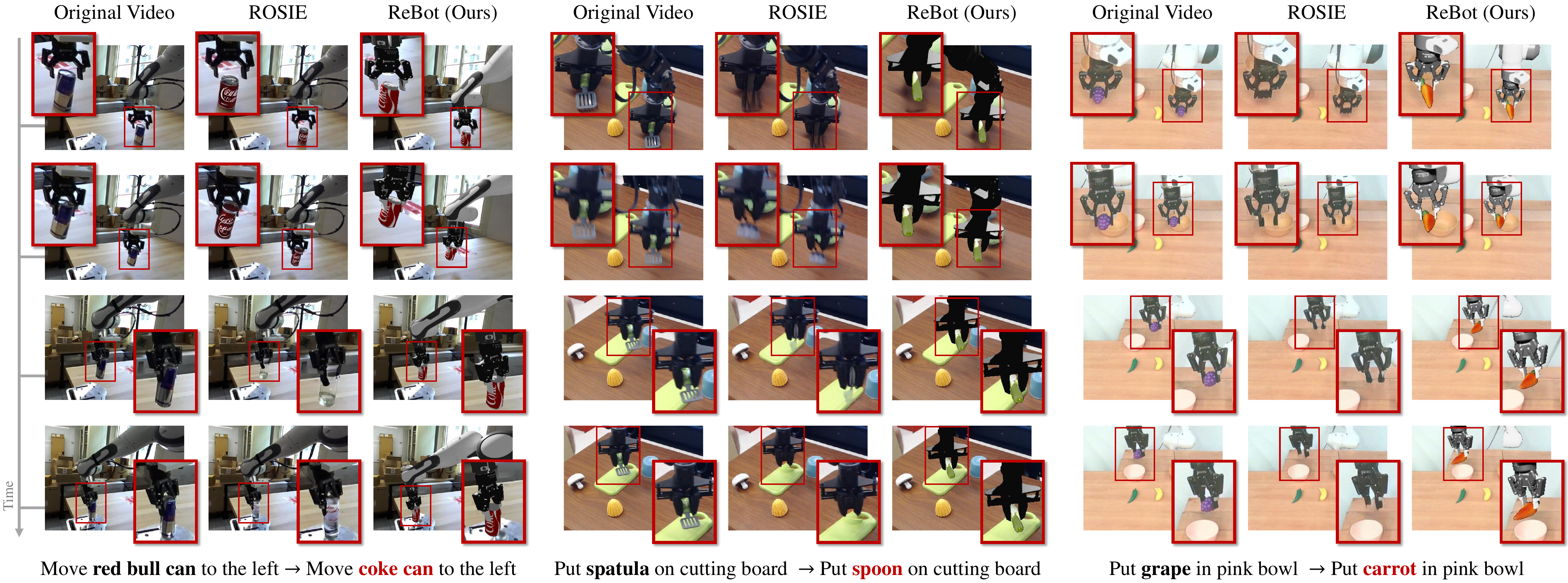}
   \caption{\textbf{Comparison of synthetic videos.} We show examples from three datasets: DROID (left), BridgeData V2 (mid), and our dataset (right). ReBot generates realistic videos with physically plausible movements and excellent temporal consistency, significantly outperforming ROSIE.}
   \label{fig:results}
   \vspace{-1em}
\end{figure*}

\noindent\textbf{Trajectory Replay.}
We reuse the real-world trajectory to diversify manipulated objects.
First, to ensure the robot can successfully reach the simulated object, we need to place it exactly where the original real object was placed.
We analyze the gripper action sequence to determine $t_{\text{start}}$ (when the gripper closes to grasp the object) and $t_{\text{end}}$ (when the gripper opens to place the object).
To estimate the object position, we acquire the gripper position at $t_{\text{start}}$ by replaying $\{\mathbf{a}_{t}\}_{t=1}^{t_\text{start}}$, and place the simulated object accordingly.
Similarly, and optionally, we place a container on the table at the gripper position at $t_{\text{end}}$.
Finally, we replay the robot trajectory using the action sequence $\{\mathbf{a}_{t}\}_{t=1}^T$, and record simulated movements $\{\mathbf{o}_{t}^{\text{sim}}\}_{t=1}^T$ for manipulating the new object.
Note that all digital twins are faithfully aligned to the real-world scene, this ensure the recorded movements remain aligned with the real-world background.

\noindent\textbf{Replay Validation.}
Notably, trajectory replay may succeed or fail in manipulating a new object, depending on the affordance compatibility between the new object and the original real-world object.
We automatically validate whether the object is successfully manipulated in each synthetic episode, and discard failed episodes by monitoring the Cartesian distance between the object and gripper from $t_{\text{start}}$ to $t_{\text{end}}$.
We present a representative example in Fig.~\ref{fig:method}, showing that despite the disparity of object shapes, real-world trajectories can be successfully reused to manipulate various objects, demonstrating the scalability of our approach.

\subsection{Real-world Background Inpainting}
\label{sec:real-world}
In this step, we prepare task-agnostic real-world background $\{\mathbf{o}_{t}^{\text{real}}\}_{t=1}^T$ for integration with simulated movements, by removing task-specific elements (\emph{i.e.}, the original real object and robot) in the original real robot video $\{\mathbf{o}_{t}\}_{t=1}^T$.

\noindent\textbf{Object and Robot Segmentation.}
We automatically segment and track the original real object and robot by using GroundedSAM2~\cite{ren2024grounded}, which combines GroundingDINO~\cite{liu2023grounding} and SAM2~\cite{ravi2024sam2segmentimages}.
More specifically, we first use GroundingDINO to identify and segment the robot using the text prompt (``robot'') on $\mathbf{o}_{t_{\text{start}}}$, as we empirically observe the best performance when the robot is most visible.
However, automatically identifying the original real object is extremely challenging, as a detailed description of its appearance, which is essential for effective text prompts, is typically unavailable in real robot datasets.
Moreover, text prompts are highly susceptible to distractors or similar instances, making them unreliable for accurately locating the manipulated object.
Fortunately, the object position at $t_{\text{start}}$ is already estimated during real-to-sim trajectory replay, now serving as a crucial cue for segmenting the real object on $\mathbf{o}_{t_{\text{start}}}$.
Using the camera pose, we project the 3D object position onto $\mathbf{o}_{t_{\text{start}}}$, providing a 2D point prompt for real object segmentation with SAM2.
After obtaining the semantic mask $\mathbf{m}_{t_{\text{start}}}$ (\emph{i.e.}, the robot and object masks at $t_{\text{start}}$), we propagate it to all video frames $\{\mathbf{o}_{t}\}_{t=1}^T$ using SAM2, generating the corresponding semantic masks $\{\mathbf{m}_{t}\}_{t=1}^T$.

\noindent\textbf{Object and Robot Removal.}
Given $\{\mathbf{o}_{t}, \mathbf{m}_{t}\}_{t=1}^T$, we eventually apply ProPainter~\cite{zhou2023propainter}, a state-of-the-art video inpainting model, to remove the original real object and robot from the original video, obtaining the task-agnostic background $\{\mathbf{o}_{t}^{\text{real}}\}_{t=1}^T$.
Notice that we also remove the real robot in this step and later use the virtual robot in our synthetic videos $\{\mathbf{o}'_t\}_{t=1}^T$. 
This ensures correct occlusions and realistic physical interactions during object manipulation.

\subsection{Sim-to-Real Video Synthesis}
\label{sec:sim-to-real}
We eventually combine simulated movements $\{\mathbf{o}_{t}^{\text{sim}}\}_{t=1}^T$ with task-agnostic real-world background $\{\mathbf{o}_{t}^{\text{real}}\}_{t=1}^T$ to build new video frames $\{\mathbf{o}'_t\}_{t=1}^{T}$.
Specifically, to obtain $\mathbf{o}'_{t}$, we extract the robot and the manipulated object from $\mathbf{o}_{t}^{\text{sim}}$,
and merge them onto $\mathbf{o}_{t}^{\text{real}}$.
We then assign a new language instruction $\mathcal{L}'$ by replacing the object (\emph{e.g.}, ``yellow mug" to ``spoon") and container (\emph{e.g.}, ``table" to ``towel") in the original instruction $\mathcal{L}$ to the ones we used during trajectory replay.
Eventually, we construct a new episode $\mathbf{\tau}'_j = \{\mathbf{o}'_t, \mathbf{a}_t, \mathcal{L}'\}_{t=1}^T$.
Note that, since we faithfully replay real-world robot trajectories, the real-world actions remain unchanged in synthetic episodes.
In our experiments (see Sec.~\ref{sec:exp}), we validate the effectiveness of our method for adapting VLA models with our synthetic dataset $\mathcal{D}' = \{\mathbf{\tau}'_j\}_{j=1}^N$.

\section{Experiments}
\label{sec:exp}
In this section, we evaluate and demonstrate that ReBot effectively produces high-fidelity synthetic robot videos (Sec.~\ref{sec:video}), and comprehensively enhances the performance of VLA models in both simulation (Sec.~\ref{sec:eval_sim}) and real-world environments (Sec.~\ref{sec:eval_real}). 

\subsection{Experimental Setups}
\noindent\textbf{Datasets.}~
For real robot datasets, we leverage tabletop pick-and-place episodes in BridgeData V2~\cite{walke2023bridgedata} and DROID~\cite{khazatsky2024droid}.
For evaluation in real-world environments in Sec.~\ref{sec:eval_real}, we collect 220 real-world episodes to build our dataset.
In the DROID dataset, we leverage two exterior videos captured from opposite sides of the robot.
For simulated objects used in real-to-sim trajectory replay, we follow~\cite{li2024evaluating, nasiriany2024robocasa} and collect kitchen assets from Objaverse~\cite{objaverse}.

\noindent\textbf{Implementation Details.}~ 
We use Isaac Sim 4.1 as our simulation environment for its excellent rendering quality and flexibility.
We implement the real-to-sim trajectory replay based on Isaac Lab~\cite{mittal2023orbit}.
We pre-build digital twins of the robots in Isaac Sim, matching the same robot platforms as per real robot datasets, \emph{i.e.}, using WidowX 250 6DOF robot arm for BridgeData V2 and Franka Panda 7DoF robot arm with Robotiq 2F-85 gripper for DROID and our dataset.
Following the official guidelines of Octo and OpenVLA, we use 100 synthetic episodes per task as the optimal data volume for finetuning.
We use four NVIDIA A6000 GPUs, using full finetuning with a batch size of 256 and a learning rate of $4 \times 10^{-5}$ for Octo, and LoRA finetuning with a batch size of 32 and a learning rate of $5 \times 10^{-4}$ for OpenVLA.

\noindent\textbf{Methods for Comparison.}~
We compare ReBot with ROSIE~\cite{yu2023scaling}, a state-of-the-art generation-based method for scaling real robot videos.
ROSIE employs image-based foundation models, using Imagen~\cite{saharia2022photorealistic} to inpaint manipulated objects directly on original real robot videos.
In contrast, ReBot introduces a novel real-to-sim-to-real scaling strategy, producing physically realistic and temporally consistent synthetic robot videos.
Since ROSIE is not open-source, we use our implementation based on the stable diffusion model~\cite{rombach2022high}.

\noindent\textbf{Evaluation with VLA Models.}~
We evaluate the effectiveness of synthetic videos for adapting VLA models to target domains.
We mainly discuss two state-of-the-art VLA models, Octo~\cite{octo_2023} and OpenVLA~\cite{kim2024openvla}, both of which are trained on large and diverse datasets involving various robotic embodiments~\cite{o2023open}.
To compare scaling methods, we evaluate three versions of each VLA model: 
1) Octo and OpenVLA (zero-shot evaluation, \emph{i.e.}, pre-trained models without finetuning), 
2) Octo+ROSIE and OpenVLA+ROSIE (finetuned with episodes from ROSIE),
and 3) Octo+ReBot and OpenVLA+ReBot (finetuned with episodes from ReBot).

\subsection{Evaluation of Video Quality}
\label{sec:video}
\begin{figure}[t]
  \centering
  \includegraphics[width=0.9\linewidth]{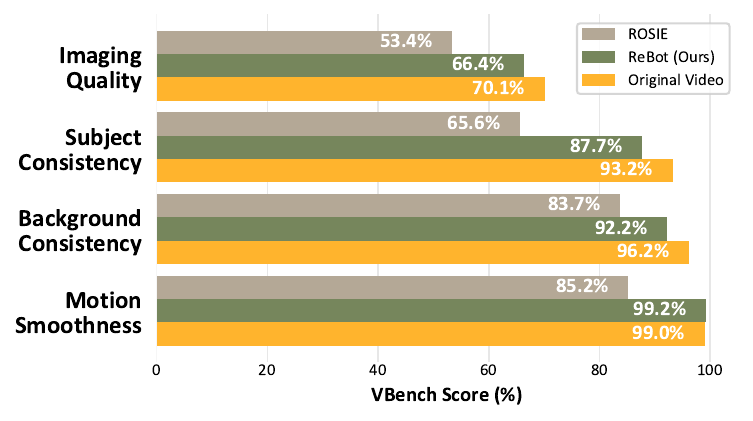}
  \vspace{-1em}
   \caption{\textbf{Quantitative comparison of generated video quality.} We report VBench scores as evaluation metrics. \alias outperforms ROSIE and achieves video quality comparable to original real-world videos.}
   \vspace{-12pt}
   \label{fig:vbench}
\end{figure}

We compare the generated video quality of ROSIE~\cite{yu2023scaling} and ReBot across three aspects: Temporal Quality, Imaging Quality, and Multi-view Consistency.
We present a qualitative comparison in Fig.~\ref{fig:results}. 
Meanwhile, as shown in Fig.~\ref{fig:vbench}, we use VBench~\cite{huang2023vbench}, a comprehensive benchmark tool for assessing video generation quality, to evaluate two key aspects across four dimensions (please refer to~\cite{huang2023vbench} for detailed definitions): 1) Temporal Quality - including Subject Consistency, Background Consistency, and Motion Smoothness; and 2) Frame-wise Quality, \emph{i.e.}, Imaging Quality.
We also evaluate original real videos for reference.

\noindent\textbf{Temporal Quality.}~
Although ROSIE offers a straightforward solution, it fails to generate temporally consistent videos, which hinders VLA models from learning stable actions.
As shown in the first example of Fig.~\ref{fig:results}, ROSIE initially generates a plausible coke can in the first two frames, but then fails to maintain consistency, producing irrelevant bottles in later frames.
This limitation is further reflected in its low subject consistency score of only $65.6\%$, as reported in Fig.~\ref{fig:vbench}.
Therefore, although observation history has been shown to enhance VLA models~\cite{brohan2022rt, octo_2023}, ROSIE remains unsuitable for improving their ability to learn from consecutive frames.
In contrast, ReBot inherently ensures excellent temporal consistency through the simulation process, achieving $99.2\%$ in motion smoothness.
Surprisingly, this even slightly outperforms real robot videos by 0.2\%, possibly because the simulation process reduces artifacts such as motion blur (see the second frame in the second example in Fig.~\ref{fig:results}).
Moreover, real-world background inpainting faithfully uses temporal context to recover the occlusions, contributing to a $92.2\%$ background consistency.
Notably, our temporal quality across all dimensions, with an average score of 93.0\%, is highly comparable to real robot videos (96.1\%), indicating that our synthetic videos achieve lifelike temporal consistency.

\noindent\textbf{Imaging Quality.}~
In Fig.~\ref{fig:results}, ROSIE struggles to generate high-quality manipulated objects, especially in the last two examples.
This issue becomes particularly evident when the new object shape potentially deviates from the original object shape.
This is because generative models tend to rely more on the inpainting mask, while paying less attention to the guidance of the text prompt.
By comparison, ReBot ensures physically plausible movements through simulation, while demonstrating excellent imaging quality in Fig.~\ref{fig:vbench}, with only a 3.7\% decrease compared to original videos, while surpassing ROSIE by 13.0\%.

\begin{figure}[t]
  \centering
   \includegraphics[width=0.9\linewidth]{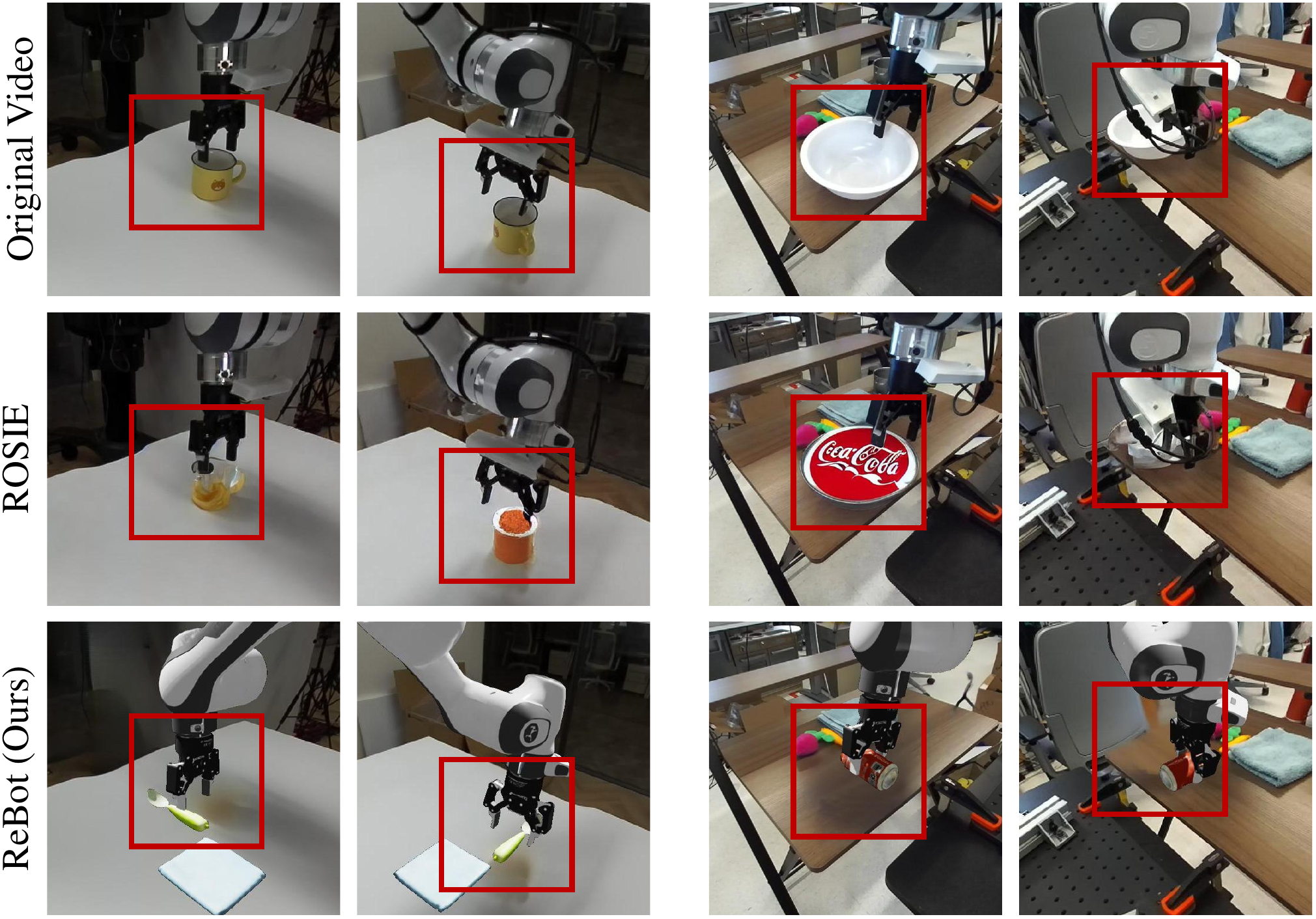}
   \vspace{-0.5em}
   \caption{\textbf{Comparisons of multi-view consistency}. We present two examples from the DROID dataset, each captured from two different camera views.
   While ROSIE lacks multi-view consistency, ReBot naturally preserves this capability inherited from 3D simulation, ensuring the same object in different camera views, as in the real world.
   }
   \label{fig:multi-view}
   \vspace{-0.5em}
\end{figure}

\noindent\textbf{Multi-view Consistency.}~
Additionally, as shown in Fig.~\ref{fig:multi-view}, \alias inherently preserves multi-view consistency across multiple camera views, since the synthetic videos are produced within a 3D environment.
Notably, this crucial attribute is uniquely achievable through our real-to-sim-to-real scaling approach.

\begin{table*}[t]
    \caption{Comparison of Evaluation Results on the WidowX Robot in SimplerEnv.}
    \small
    \centering
    \begin{tabular} {>{\centering\arraybackslash}p{3.3cm} | >{\centering\arraybackslash}p{0.9cm}  >{\centering\arraybackslash}p{0.9cm} | >{\centering\arraybackslash}p{0.9cm} >{\centering\arraybackslash}p{0.9cm} | >{\centering\arraybackslash}p{0.9cm} >{\centering\arraybackslash}p{0.9cm} | >{\centering\arraybackslash}p{0.9cm} >{\centering\arraybackslash}p{0.9cm} | >{\centering\arraybackslash}p{0.9cm} >{\centering\arraybackslash}p{0.9cm}}
    
    \toprule
    \multirow{3}{*}{\textbf{Model}}  &  \multicolumn{2}{c}{\textbf{\footnotesize \makecell{ Put spoon \\ on towel}}} & \multicolumn{2}{|c}{\textbf{\footnotesize \makecell { Put carrot \\ on plate}}} & \multicolumn{2}{|c}{\textbf{\footnotesize \makecell{Stack green cube \\ on yellow cube}}} & \multicolumn{2}{|c}{\textbf{\footnotesize \makecell{Put eggplant \\ in basket}}} & \multicolumn{2}{|c}{\textbf{Average}}\\
    \cmidrule{2-11}
    & \footnotesize Grasp & \footnotesize Success & \footnotesize Grasp & \footnotesize Success & \footnotesize Grasp & \footnotesize Success & \footnotesize Grasp & \footnotesize Success & \footnotesize Grasp & \footnotesize Success \\
    
    \midrule
    Octo~\cite{octo_2023}         & 34.7\% & 12.5\% & \textbf{52.8\%} & 8.3\% & 31.9\% & 0.0\% & \textbf{66.7\%} & \textbf{43.1\%} & 46.5\% & 16.0\% \\
    Octo+ROSIE~\cite{yu2023scaling}     & 20.8\% & 2.8\% & 27.8\% & 0.0\% & 18.1\% & 0.0\% & 22.3\% & 0.0\% & 22.3\% & 0.7\% \\
    Octo+ReBot (Ours) & \textbf{61.1\%} & \textbf{54.2\%} & 41.1\% & \textbf{22.0\%} & \textbf{63.9\%} & \textbf{4.2\%} & 52.8\% & 12.5\% & \textbf{54.7\%} & \textbf{23.2\%} \\
    
    \midrule
    OpenVLA~\cite{kim2024openvla} & 4.2\% & 0.0\% & 33.3\% & 0.0\% & 12.5\% & 0.0\% & 8.3\% & 4.2\% & 14.6\% & 1.1\% \\
    OpenVLA+ROSIE~\cite{yu2023scaling} & 12.5\% & 0.0\% & 41.7\% & 0.0\% & 50.0\% & 0.0\% & 20.8\% & 0.0\% & 31.3\% & 0.0\% \\
    OpenVLA+ReBot (Ours) & \textbf{58.3\%} & \textbf{20.8\%} & \textbf{45.8\%} & \textbf{12.5\%} & \textbf{66.7\%} & \textbf{4.2\%} & \textbf{66.7\%} & \textbf{54.2\%} & \textbf{59.4\%} & \textbf{22.9\%} \\
    
    \bottomrule
    \end{tabular}
    \label{tab:results}
\end{table*}

\begin{figure*}[h]
  \centering
   \includegraphics[width=1.0\linewidth]{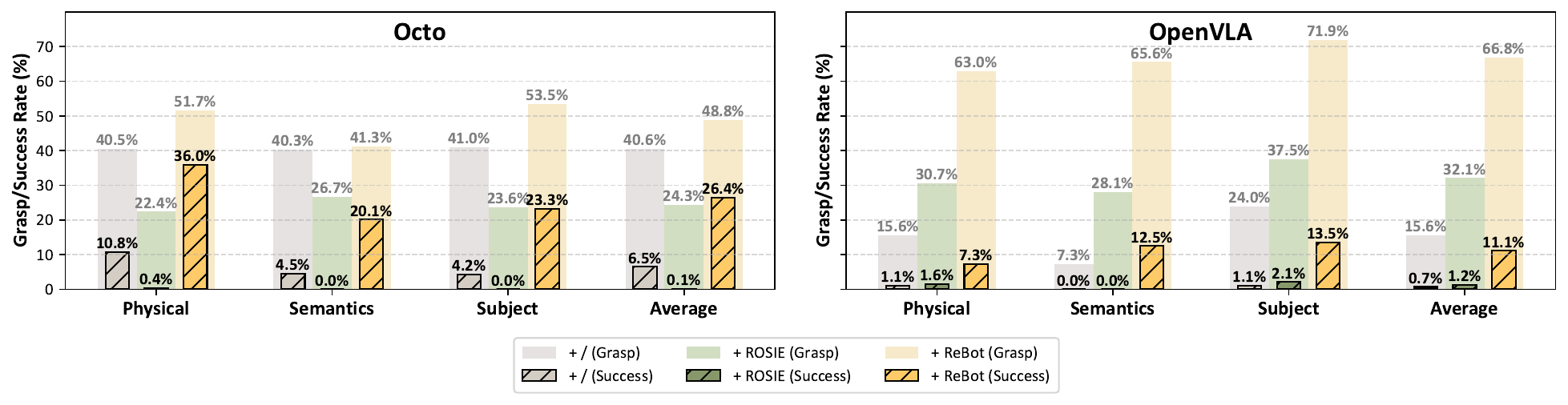}
   \vspace{-2em}
   \caption{\textbf{Evaluation of generalization performance.} ReBot improves the generalization performance of Octo (left) and OpenVLA (right) across all three generalization types (physical, semantics, and subject) on WidowX Robot in SimplerEnv.}
   \vspace{-1em}
   \label{fig:generalization}
\end{figure*}

\begin{figure}[h]
  \centering
   \includegraphics[width=1.0\linewidth]{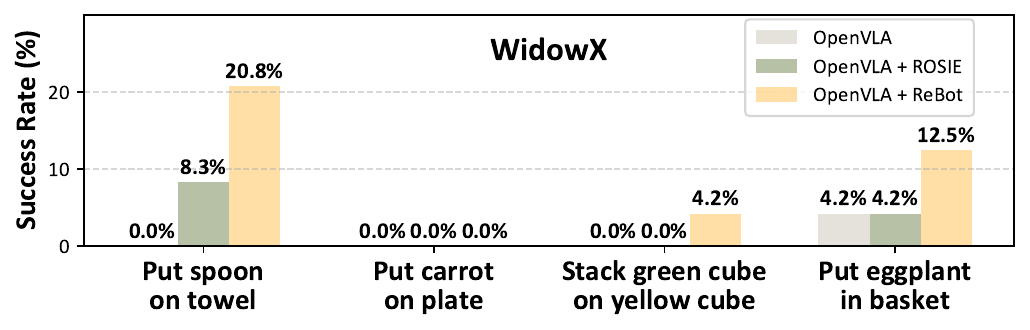}
  \includegraphics[width=1.0\linewidth]{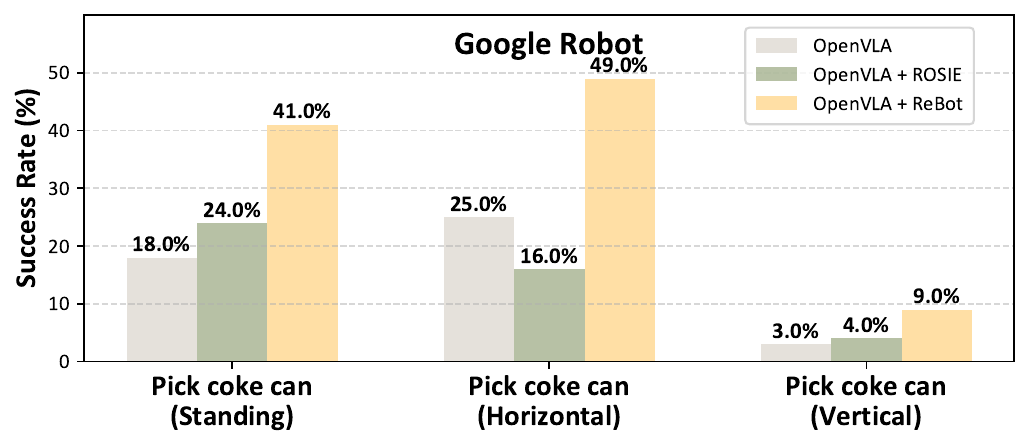}
  \vspace{-2em}
   \caption{\textbf{Evaluation of cross-embodiment performance. } ReBot enhances the cross-embodiment performance of OpenVLA on the WidowX robot (top) and Google Robot (bottom) in SimplerEnv. }
   \label{fig:cross}
   \vspace{-1.5em}
\end{figure}

\subsection{Evaluation in Simulation Environment}
\label{sec:eval_sim}
We first evaluate VLA models and their two finetuned versions (``+ROSIE" and ``+ReBot") in SimplerEnv~\cite{li2024evaluating}.
For fair comparisons, we use ROSIE and ReBot to scale the same data volume exclusively for evaluation tasks (\emph{i.e.}, 100 episodes per task), adapting VLA models to the same target domain.
We demonstrate that ReBot effectively improves VLA performance across three key aspects:
1) In-domain Performance: Direct evaluation on the given tasks;
2) Generalization Performance (following~\cite{kim2024openvla, zhang2024grape}): Evaluating variations of in-domain tasks across unseen object sizes (physical), unseen instructions (semantics), and unseen objects (subject);
3) Cross-embodiment Performance: Evaluating on one embodiment while finetuning on another.

\noindent\textbf{In-domain Performance.}~
In Table~\ref{tab:results}, we report the grasp rates (percentage of successful object grasps during the task) and success rates (percentage of completed tasks) for the four SimplerEnv tasks on the WidowX robot.
When used out-of-the-box, both Octo and OpenVLA struggle to report decent performance on most tasks.
Particularly, OpenVLA entirely fails on challenging tasks, showing 0.0\% success rates (\emph{e.g.}, stack green cube on yellow cube).
This demonstrates their poor performance in the target domain without data scaling, despite extensive training on SOTA-scale datasets~\cite{padalkar2023open}.
Meanwhile, ROSIE performs poorly across most tasks with 0.0\% success rates, as it fails to generate realistic manipulated objects and, more importantly, lacks temporal consistency.
This limitation is particularly problematic for Octo, which relies on observation history with two consecutive frames.
In contrast, ReBot achieves the best performance improving all models, increasing the average success rate by 7.2\% for Octo and 21.8\% for OpenVLA.
Notably, ReBot boosts the average grasp rate from 14.6\% to 59.4\% on OpenVLA, further demonstrating its effectiveness.
These results highlight that both VLA models benefit greatly from ReBot because of its temporal consistent and physically realistic synthetic videos.

\noindent\textbf{Generalization Performance.}~
While current VLA models often face generalization challenges, we further validate ReBot as an effective scaling solution for enhancing their generalization performance.
As shown in Fig.~\ref{fig:generalization}, ROSIE remains ineffective on Octo, while ReBot consistently improves both Octo and OpenVLA across all three generalization types.
Specifically, ReBot increases the average success rate from 6.5\% to 26.4\% on Octo.
On the other hand, although OpenVLA faces greater challenges in SimplerEnv, it benefits significantly from ReBot, with the average grasp rate rising from 15.6\% to 66.8\%, and the average success rate increasing from 0.7\% to 11.1\%.
These results further confirm the effectiveness of ReBot in improving the generalization performance of VLA models.

\begin{table*}
    \caption{Comparison of evaluation results on the Franka Panda robot in the real world environment.}
    \small
    \centering
    \begin{tabular} {>{\centering\arraybackslash}p{3.3cm} | > {\centering\arraybackslash}p{0.8cm}  >{\centering\arraybackslash}p{0.8cm} | >{\centering\arraybackslash}p{0.8cm} >{\centering\arraybackslash}p{0.8cm} | >{\centering\arraybackslash}p{0.8cm} >{\centering\arraybackslash}p{0.8cm} | >{\centering\arraybackslash}p{0.8cm} >{\centering\arraybackslash}p{0.8cm} | >{\centering\arraybackslash}p{0.8cm} >{\centering\arraybackslash}p{0.8cm}}
    
    \toprule
    \multirow{3}{*}{\textbf{Model}}  &  \multicolumn{2}{c}{\textbf{\makecell{Put carrot \\ in blue plate}}} & \multicolumn{2}{|c}{\textbf{\makecell {Put grape \\ in yellow plate}}} & \multicolumn{2}{|c}{\textbf{\makecell{Put fanta can \\ in blue plate}}} & \multicolumn{2}{|c}{\textbf{\makecell{Put black cube \\ in yellow plate}}} & \multicolumn{2}{|c}{\textbf{Average}}\\
    \cmidrule{2-11}
    & \footnotesize Grasp & \footnotesize Success & \footnotesize Grasp & \footnotesize Success & \footnotesize Grasp & \footnotesize Success & \footnotesize Grasp & \footnotesize Success & \footnotesize Grasp & \footnotesize Success \\
    
    \midrule
    Octo~\cite{octo_2023}         & 0\% & 0\% & 30\% & 20\% & 10\% & 0\% & 20\% & 10\% & 15\% & 8\% \\
    Octo+ROSIE~\cite{yu2023scaling}   & 30\% & 20\% & 0\% & 0\% & 20\% & 20\% & 10\% & 0\% & 15\% & 10\% \\
    Octo+ReBot (Ours)  & \textbf{40\%} & \textbf{20\%} & \textbf{40\%} & \textbf{30\%} & \textbf{30\%} & \textbf{20\%} & \textbf{30\%} & \textbf{30\%} & \textbf{35\%} & \textbf{25\%} \\
    
    \midrule
    OpenVLA~\cite{kim2024openvla}         & 30\% & 20\% & 30\% & 20\% & \textbf{60\%} & 30\% & 40\% & 30\% & 40\% & 25\% \\
    OpenVLA+ROSIE~\cite{yu2023scaling}         & 10\% & 0\% & 10\% & 0\% & 30\% & 10\% & 20\% & 10\% & 18\% & 5\% \\
    OpenVLA+ReBot (Ours) & \textbf{40\%} & \textbf{40\%} & \textbf{50\%} & \textbf{40\%} & 50\% & \textbf{50\%} & \textbf{60\%} & \textbf{50\%} & \textbf{50\%} & \textbf{45\%} \\
    
    \bottomrule
    \end{tabular}
    \label{tab:real}
\includegraphics[width=0.8\linewidth]{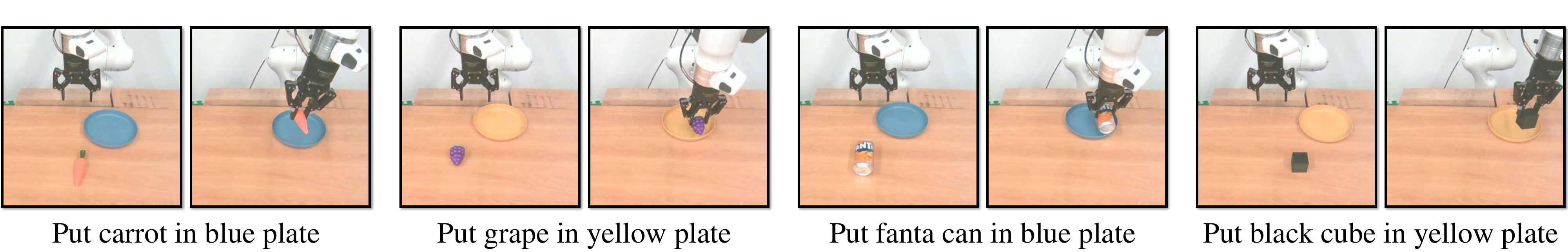}
\vspace{-1em}
\end{table*}

\noindent\textbf{Cross-embodiment Performance.}~
We also investigate whether ReBot can enhance the cross-embodiment performance of VLA models.
Specifically, we use ROSIE and ReBot to scale the DROID dataset for the Franka Panda robot, then finetune OpenVLA and evaluate its performance on the WidowX robot and Google Robot in SimplerEnv.
We report the success rates in Fig.~\ref{fig:cross}.
On the WidowX robot, while ROSIE only provides a marginal increase in the average success rate from 1.4\% to 3.1\%, ReBot achieves a substantial boost to 12.5\%. 
For the ``pick coke can" task with varying object poses on Google Robot, ReBot demonstrates consistent improvements across all poses, whereas ROSIE fails to achieve such robustness.
This highlights that ReBot enables OpenVLA to learn more precise and adaptable manipulation strategies across diverse object poses.
Notably, ReBot consistently improves the performance of OpenVLA despite scaling for a different embodiment, demonstrating its ability to enhance cross-embodiment performance.

\subsection{Evaluation in Real-world Environment}
\label{sec:eval_real}
In real-world experiments, we demonstrate that ReBot consistently enhances the effectiveness of VLA models, delivering superior performance over ROSIE.
As shown in Tab.~\ref{tab:real}, we leverage both ROSIE and ReBot to scale our real robot dataset for four evaluation tasks (see examples at the bottom of Tab.~\ref{tab:real}), and compare the performance of their finetuned VLA models.
To ensure better adaptation to our real-world scene, we also incorporate our real robot dataset (\emph{i.e.}, 220 real-world episodes) during the finetuning process for all models.
We conduct 10 trials per task, and report both the grasp rate and success rate as evaluation metrics.
While ROSIE provides a marginal improvement, increasing the average success rate of Octo from 8\% to 10\%, it fails entirely on some tasks (\emph{e.g.}, put the grape in yellow plate), and does not show meaningful enhancement for OpenVLA.
In contrast, ReBot consistently achieves substantial performance gains across diverse tasks, improving the average success rates of Octo by 17\% and OpenVLA by 20\%.
Notably, for challenging tasks where Octo initially has a 0\% grasp rate and success rate (\emph{e.g.}, put carrot in blue plate), ReBot boosts the grasping rate to 40\% and the success rate to 20\%, highlighting its robust effectiveness in real-world applications.

\section{Conclusion and Discussion}
We propose \alias, a novel real-to-sim-to-real approach for scaling real robot datasets and adapting VLA models to target domains.
ReBot replays real-world robot trajectories in simulation to diversify manipulated objects, and integrates the simulated movements with inpainted real-world background to synthesize physically realistic and temporally consistent robot videos.
\alias achieves excellent video generation quality, with a VBench temporal consistency score of 93.0\% and imaging quality score of 66.4\%, which are comparable to 96.1\% and 70.1\% for real robot videos.
In SimplerEnv with the WidowX robot, \alias improved the in-domain performance of Octo by 7.2\% and OpenVLA by 21.8\%, and enhanced generalization performance by 19.9\% and 9.4\%, respectively.
In a real-world environment with a physical Franka Panda, ReBot increased the success rates of Octo by 17\% and OpenVLA by 20\%.

We hope \alias could serve as a valuable asset and inspire future research on real-to-sim-to-real for robot learning.
It opens several exciting avenues for future exploration.
For example, extending \alias to diverse data settings (\emph{e.g.}, varying camera setups and robots) could potentially benefit cross-embodiment learning. 
Additionally, exploring more challenging scenarios beyond tabletop manipulation is also interesting with potential broader real-world applications.
We take these directions for future work.

{
    \footnotesize
    \bibliographystyle{ieeetran}
    \bibliography{main}

\begin{thebibliography}{10}
\providecommand{\url}[1]{#1}
\csname url@rmstyle\endcsname
\providecommand{\newblock}{\relax}
\providecommand{\bibinfo}[2]{#2}
\providecommand\BIBentrySTDinterwordspacing{\spaceskip=0pt\relax}
\providecommand\BIBentryALTinterwordstretchfactor{4}
\providecommand\BIBentryALTinterwordspacing{\spaceskip=\fontdimen2\font plus
\BIBentryALTinterwordstretchfactor\fontdimen3\font minus \fontdimen4\font\relax}
\providecommand\BIBforeignlanguage[2]{{%
\expandafter\ifx\csname l@#1\endcsname\relax
\typeout{** WARNING: IEEEtran.bst: No hyphenation pattern has been}%
\typeout{** loaded for the language `#1'. Using the pattern for}%
\typeout{** the default language instead.}%
\else
\language=\csname l@#1\endcsname
\fi
#2}}

\bibitem{brohan2022rt}
A.~Brohan, N.~Brown, J.~Carbajal, Y.~Chebotar, J.~Dabis, C.~Finn, K.~Gopalakrishnan, K.~Hausman, A.~Herzog, J.~Hsu, \emph{et~al.}, ``Rt-1: Robotics transformer for real-world control at scale,'' \emph{arXiv preprint arXiv:2212.06817}, 2022.

\bibitem{brohan2023rt}
A.~Brohan, N.~Brown, J.~Carbajal, Y.~Chebotar, X.~Chen, K.~Choromanski, T.~Ding, D.~Driess, A.~Dubey, C.~Finn, \emph{et~al.}, ``Rt-2: Vision-language-action models transfer web knowledge to robotic control,'' \emph{arXiv preprint arXiv:2307.15818}, 2023.

\bibitem{padalkar2023open}
A.~Padalkar, A.~Pooley, A.~Jain, A.~Bewley, A.~Herzog, A.~Irpan, A.~Khazatsky, A.~Rai, A.~Singh, A.~Brohan, \emph{et~al.}, ``Open x-embodiment: Robotic learning datasets and rt-x models,'' \emph{arXiv preprint arXiv:2310.08864}, 2023.

\bibitem{o2023open}
A.~O'Neill, A.~Rehman, A.~Gupta, A.~Maddukuri, A.~Gupta, A.~Padalkar, A.~Lee, A.~Pooley, A.~Gupta, A.~Mandlekar, \emph{et~al.}, ``Open x-embodiment: Robotic learning datasets and rt-x models,'' \emph{arXiv preprint arXiv:2310.08864}, 2023.

\bibitem{khazatsky2024droid}
A.~Khazatsky, K.~Pertsch, S.~Nair, A.~Balakrishna, S.~Dasari, S.~Karamcheti, S.~Nasiriany, M.~K. Srirama, L.~Y. Chen, K.~Ellis, \emph{et~al.}, ``Droid: A large-scale in-the-wild robot manipulation dataset,'' \emph{arXiv preprint arXiv:2403.12945}, 2024.

\bibitem{kolve2017ai2}
E.~Kolve, R.~Mottaghi, W.~Han, E.~VanderBilt, L.~Weihs, A.~Herrasti, M.~Deitke, K.~Ehsani, D.~Gordon, Y.~Zhu, \emph{et~al.}, ``Ai2-thor: An interactive 3d environment for visual ai,'' \emph{arXiv preprint arXiv:1712.05474}, 2017.

\bibitem{mu2021maniskill}
T.~Mu, Z.~Ling, F.~Xiang, D.~Yang, X.~Li, S.~Tao, Z.~Huang, Z.~Jia, and H.~Su, ``Maniskill: Generalizable manipulation skill benchmark with large-scale demonstrations,'' \emph{arXiv preprint arXiv:2107.14483}, 2021.

\bibitem{gu2023maniskill2}
J.~Gu, F.~Xiang, X.~Li, Z.~Ling, X.~Liu, T.~Mu, Y.~Tang, S.~Tao, X.~Wei, Y.~Yao, \emph{et~al.}, ``Maniskill2: A unified benchmark for generalizable manipulation skills,'' \emph{arXiv preprint arXiv:2302.04659}, 2023.

\bibitem{wang2023robogen}
Y.~Wang, Z.~Xian, F.~Chen, T.-H. Wang, Y.~Wang, K.~Fragkiadaki, Z.~Erickson, D.~Held, and C.~Gan, ``Robogen: Towards unleashing infinite data for automated robot learning via generative simulation,'' \emph{arXiv preprint arXiv:2311.01455}, 2023.

\bibitem{liu2023libero}
B.~Liu, Y.~Zhu, C.~Gao, Y.~Feng, Q.~Liu, Y.~Zhu, and P.~Stone, ``Libero: Benchmarking knowledge transfer for lifelong robot learning,'' \emph{arXiv preprint arXiv:2306.03310}, 2023.

\bibitem{nasiriany2024robocasa}
S.~Nasiriany, A.~Maddukuri, L.~Zhang, A.~Parikh, A.~Lo, A.~Joshi, A.~Mandlekar, and Y.~Zhu, ``Robocasa: Large-scale simulation of everyday tasks for generalist robots,'' \emph{arXiv preprint arXiv:2406.02523}, 2024.

\bibitem{zhao2020sim}
W.~Zhao, J.~P. Queralta, and T.~Westerlund, ``Sim-to-real transfer in deep reinforcement learning for robotics: a survey,'' in \emph{2020 IEEE symposium series on computational intelligence (SSCI)}.\hskip 1em plus 0.5em minus 0.4em\relax IEEE, 2020, pp. 737--744.

\bibitem{muratore2022robot}
F.~Muratore, F.~Ramos, G.~Turk, W.~Yu, M.~Gienger, and J.~Peters, ``Robot learning from randomized simulations: A review,'' \emph{Frontiers in Robotics and AI}, vol.~9, p. 799893, 2022.

\bibitem{mandi2022cacti}
Z.~Mandi, H.~Bharadhwaj, V.~Moens, S.~Song, A.~Rajeswaran, and V.~Kumar, ``Cacti: A framework for scalable multi-task multi-scene visual imitation learning,'' \emph{arXiv preprint arXiv:2212.05711}, 2022.

\bibitem{zhou2024robodreamer}
S.~Zhou, Y.~Du, J.~Chen, Y.~Li, D.-Y. Yeung, and C.~Gan, ``Robodreamer: Learning compositional world models for robot imagination,'' \emph{arXiv preprint arXiv:2404.12377}, 2024.

\bibitem{du2024learning}
Y.~Du, S.~Yang, B.~Dai, H.~Dai, O.~Nachum, J.~Tenenbaum, D.~Schuurmans, and P.~Abbeel, ``Learning universal policies via text-guided video generation,'' \emph{Advances in Neural Information Processing Systems}, vol.~36, 2024.

\bibitem{chen2023genaug}
Z.~Chen, S.~Kiami, A.~Gupta, and V.~Kumar, ``Genaug: Retargeting behaviors to unseen situations via generative augmentation,'' \emph{arXiv preprint arXiv:2302.06671}, 2023.

\bibitem{yu2023scaling}
T.~Yu, T.~Xiao, A.~Stone, J.~Tompson, A.~Brohan, S.~Wang, J.~Singh, C.~Tan, J.~Peralta, B.~Ichter, \emph{et~al.}, ``Scaling robot learning with semantically imagined experience,'' \emph{arXiv preprint arXiv:2302.11550}, 2023.

\bibitem{chen2024roviaug}
L.~Y. Chen, C.~Xu, K.~Dharmarajan, M.~Z. Irshad, R.~Cheng, K.~Keutzer, M.~Tomizuka, Q.~Vuong, and K.~Goldberg, ``Rovi-aug: Robot and viewpoint augmentation for cross-embodiment robot learning,'' in \emph{Conference on Robot Learning (CoRL)}, Munich, Germany, 2024.

\bibitem{ren2024grounded}
T.~Ren, S.~Liu, A.~Zeng, J.~Lin, K.~Li, H.~Cao, J.~Chen, X.~Huang, Y.~Chen, F.~Yan, Z.~Zeng, H.~Zhang, F.~Li, J.~Yang, H.~Li, Q.~Jiang, and L.~Zhang, ``Grounded sam: Assembling open-world models for diverse visual tasks,'' 2024.

\bibitem{zhou2023propainter}
S.~Zhou, C.~Li, K.~C. Chan, and C.~C. Loy, ``{ProPainter}: Improving propagation and transformer for video inpainting,'' in \emph{Proceedings of IEEE International Conference on Computer Vision (ICCV)}, 2023.

\bibitem{ravichandar2020recent}
H.~Ravichandar, A.~S. Polydoros, S.~Chernova, and A.~Billard, ``Recent advances in robot learning from demonstration,'' \emph{Annual review of control, robotics, and autonomous systems}, vol.~3, no.~1, pp. 297--330, 2020.

\bibitem{yang2024enhancing}
Y.~Yang, L.~Chen, Z.~Zaidi, S.~van Waveren, A.~Krishna, and M.~Gombolay, ``Enhancing safety in learning from demonstration algorithms via control barrier function shielding,'' in \emph{Proceedings of the 2024 ACM/IEEE International Conference on Human-Robot Interaction}, 2024, pp. 820--829.

\bibitem{mandlekar2019scaling}
A.~Mandlekar, J.~Booher, M.~Spero, A.~Tung, A.~Gupta, Y.~Zhu, A.~Garg, S.~Savarese, and L.~Fei-Fei, ``Scaling robot supervision to hundreds of hours with roboturk: Robotic manipulation dataset through human reasoning and dexterity,'' in \emph{2019 IEEE/RSJ International Conference on Intelligent Robots and Systems (IROS)}.\hskip 1em plus 0.5em minus 0.4em\relax IEEE, 2019, pp. 1048--1055.

\bibitem{ebert2021bridge}
F.~Ebert, Y.~Yang, K.~Schmeckpeper, B.~Bucher, G.~Georgakis, K.~Daniilidis, C.~Finn, and S.~Levine, ``Bridge data: Boosting generalization of robotic skills with cross-domain datasets,'' \emph{arXiv preprint arXiv:2109.13396}, 2021.

\bibitem{jang2022bc}
E.~Jang, A.~Irpan, M.~Khansari, D.~Kappler, F.~Ebert, C.~Lynch, S.~Levine, and C.~Finn, ``Bc-z: Zero-shot task generalization with robotic imitation learning,'' in \emph{Conference on Robot Learning}.\hskip 1em plus 0.5em minus 0.4em\relax PMLR, 2022, pp. 991--1002.

\bibitem{whitney2019comparing}
D.~Whitney, E.~Rosen, E.~Phillips, G.~Konidaris, and S.~Tellex, ``Comparing robot grasping teleoperation across desktop and virtual reality with ros reality,'' in \emph{Robotics Research: The 18th International Symposium ISRR}.\hskip 1em plus 0.5em minus 0.4em\relax Springer, 2019, pp. 335--350.

\bibitem{yang2024arcade}
Y.~Yang, B.~Ikeda, G.~Bertasius, and D.~Szafir, ``Arcade: Scalable demonstration collection and generation via augmented reality for imitation learning,'' in \emph{2024 IEEE/RSJ International Conference on Intelligent Robots and Systems (IROS)}.\hskip 1em plus 0.5em minus 0.4em\relax IEEE, 2024, pp. 2855--2861.

\bibitem{octo_2023}
{Octo Model Team}, D.~Ghosh, H.~Walke, K.~Pertsch, K.~Black, O.~Mees, S.~Dasari, J.~Hejna, C.~Xu, J.~Luo, T.~Kreiman, Y.~Tan, L.~Y. Chen, P.~Sanketi, Q.~Vuong, T.~Xiao, D.~Sadigh, C.~Finn, and S.~Levine, ``Octo: An open-source generalist robot policy,'' in \emph{Proceedings of Robotics: Science and Systems}, Delft, Netherlands, 2024.

\bibitem{kim2024openvla}
M.~J. Kim, K.~Pertsch, S.~Karamcheti, T.~Xiao, A.~Balakrishna, S.~Nair, R.~Rafailov, E.~Foster, G.~Lam, P.~Sanketi, \emph{et~al.}, ``Openvla: An open-source vision-language-action model,'' \emph{arXiv preprint arXiv:2406.09246}, 2024.

\bibitem{savva2019habitat}
M.~Savva, A.~Kadian, O.~Maksymets, Y.~Zhao, E.~Wijmans, B.~Jain, J.~Straub, J.~Liu, V.~Koltun, J.~Malik, \emph{et~al.}, ``Habitat: A platform for embodied ai research,'' in \emph{Proceedings of the IEEE/CVF international conference on computer vision}, 2019, pp. 9339--9347.

\bibitem{shridhar2020alfred}
M.~Shridhar, J.~Thomason, D.~Gordon, Y.~Bisk, W.~Han, R.~Mottaghi, L.~Zettlemoyer, and D.~Fox, ``Alfred: A benchmark for interpreting grounded instructions for everyday tasks,'' in \emph{Proceedings of the IEEE/CVF conference on computer vision and pattern recognition}, 2020, pp. 10\,740--10\,749.

\bibitem{xiang2020sapien}
F.~Xiang, Y.~Qin, K.~Mo, Y.~Xia, H.~Zhu, F.~Liu, M.~Liu, H.~Jiang, Y.~Yuan, H.~Wang, \emph{et~al.}, ``Sapien: A simulated part-based interactive environment,'' in \emph{Proceedings of the IEEE/CVF conference on computer vision and pattern recognition}, 2020, pp. 11\,097--11\,107.

\bibitem{mittal2023orbit}
M.~Mittal, C.~Yu, Q.~Yu, J.~Liu, N.~Rudin, D.~Hoeller, J.~L. Yuan, R.~Singh, Y.~Guo, H.~Mazhar, \emph{et~al.}, ``Orbit: A unified simulation framework for interactive robot learning environments,'' \emph{IEEE Robotics and Automation Letters}, vol.~8, no.~6, pp. 3740--3747, 2023.

\bibitem{wang2023real2sim2real}
L.~Wang, R.~Guo, Q.~Vuong, Y.~Qin, H.~Su, and H.~Christensen, ``A real2sim2real method for robust object grasping with neural surface reconstruction,'' in \emph{2023 IEEE 19th International Conference on Automation Science and Engineering (CASE)}.\hskip 1em plus 0.5em minus 0.4em\relax IEEE, 2023, pp. 1--8.

\bibitem{torne2024reconciling}
M.~Torne, A.~Simeonov, Z.~Li, A.~Chan, T.~Chen, A.~Gupta, and P.~Agrawal, ``Reconciling reality through simulation: A real-to-sim-to-real approach for robust manipulation,'' \emph{arXiv preprint arXiv:2403.03949}, 2024.

\bibitem{mu2024robotwin}
Y.~Mu, T.~Chen, S.~Peng, Z.~Chen, Z.~Gao, Y.~Zou, L.~Lin, Z.~Xie, and P.~Luo, ``Robotwin: Dual-arm robot benchmark with generative digital twins (early version),'' \emph{arXiv preprint arXiv:2409.02920}, 2024.

\bibitem{li2024robogsimreal2sim2realroboticgaussian}
\BIBentryALTinterwordspacing
X.~Li, J.~Li, Z.~Zhang, R.~Zhang, F.~Jia, T.~Wang, H.~Fan, K.-K. Tseng, and R.~Wang, ``Robogsim: A real2sim2real robotic gaussian splatting simulator,'' 2024. [Online]. Available: \url{https://arxiv.org/abs/2411.11839}
\BIBentrySTDinterwordspacing

\bibitem{li2024evaluating}
X.~Li, K.~Hsu, J.~Gu, K.~Pertsch, O.~Mees, H.~R. Walke, C.~Fu, I.~Lunawat, I.~Sieh, S.~Kirmani, \emph{et~al.}, ``Evaluating real-world robot manipulation policies in simulation,'' \emph{arXiv preprint arXiv:2405.05941}, 2024.

\bibitem{liu2023grounding}
S.~Liu, Z.~Zeng, T.~Ren, F.~Li, H.~Zhang, J.~Yang, C.~Li, J.~Yang, H.~Su, J.~Zhu, \emph{et~al.}, ``Grounding dino: Marrying dino with grounded pre-training for open-set object detection,'' \emph{arXiv preprint arXiv:2303.05499}, 2023.

\bibitem{ravi2024sam2segmentimages}
\BIBentryALTinterwordspacing
N.~Ravi, V.~Gabeur, Y.-T. Hu, R.~Hu, C.~Ryali, T.~Ma, H.~Khedr, R.~Rädle, C.~Rolland, L.~Gustafson, E.~Mintun, J.~Pan, K.~V. Alwala, N.~Carion, C.-Y. Wu, R.~Girshick, P.~Dollár, and C.~Feichtenhofer, ``Sam 2: Segment anything in images and videos,'' 2024. [Online]. Available: \url{https://arxiv.org/abs/2408.00714}
\BIBentrySTDinterwordspacing

\bibitem{walke2023bridgedata}
H.~Walke, K.~Black, A.~Lee, M.~J. Kim, M.~Du, C.~Zheng, T.~Zhao, P.~Hansen-Estruch, Q.~Vuong, A.~He, V.~Myers, K.~Fang, C.~Finn, and S.~Levine, ``Bridgedata v2: A dataset for robot learning at scale,'' in \emph{Conference on Robot Learning (CoRL)}, 2023.

\bibitem{objaverse}
M.~Deitke, D.~Schwenk, J.~Salvador, L.~Weihs, O.~Michel, E.~VanderBilt, L.~Schmidt, K.~Ehsani, A.~Kembhavi, and A.~Farhadi, ``Objaverse: A universe of annotated 3d objects,'' \emph{arXiv preprint arXiv:2212.08051}, 2022.

\bibitem{saharia2022photorealistic}
C.~Saharia, W.~Chan, S.~Saxena, L.~Li, J.~Whang, E.~L. Denton, K.~Ghasemipour, R.~Gontijo~Lopes, B.~Karagol~Ayan, T.~Salimans, \emph{et~al.}, ``Photorealistic text-to-image diffusion models with deep language understanding,'' \emph{Advances in neural information processing systems}, vol.~35, pp. 36\,479--36\,494, 2022.

\bibitem{rombach2022high}
R.~Rombach, A.~Blattmann, D.~Lorenz, P.~Esser, and B.~Ommer, ``High-resolution image synthesis with latent diffusion models,'' in \emph{Proceedings of the IEEE/CVF conference on computer vision and pattern recognition}, 2022, pp. 10\,684--10\,695.

\bibitem{huang2023vbench}
Z.~Huang, Y.~He, J.~Yu, F.~Zhang, C.~Si, Y.~Jiang, Y.~Zhang, T.~Wu, Q.~Jin, N.~Chanpaisit, Y.~Wang, X.~Chen, L.~Wang, D.~Lin, Y.~Qiao, and Z.~Liu, ``{VBench}: Comprehensive benchmark suite for video generative models,'' in \emph{Proceedings of the IEEE/CVF Conference on Computer Vision and Pattern Recognition}, 2024.

\bibitem{zhang2024grape}
Z.~Zhang, K.~Zheng, Z.~Chen, J.~Jang, Y.~Li, C.~Wang, M.~Ding, D.~Fox, and H.~Yao, ``Grape: Generalizing robot policy via preference alignment,'' \emph{arXiv preprint arXiv:2411.19309}, 2024.

\end{thebibliography}
}

\end{document}